\documentclass[11pt,a4paper]{article}
\usepackage[table]{xcolor}
\usepackage[hyperref]{eacl2021}
\usepackage{times}
\usepackage{latexsym}

\usepackage{microtype}

\aclfinalcopy % Uncomment this line for the final submission
\usepackage[]{booktabs}
\usepackage{graphicx}
\usepackage{collcell}
\usepackage{multirow}
\usepackage{pgf}

\def\colorModel{hsb} %You can use rgb or hsb

\newcommand\ColCell[1]{
  \pgfmathparse{#1<5000?1:0}  %Threshold for changing the font color into the cells
    \ifnum\pgfmathresult=0\relax\color{white}\fi
  \pgfmathsetmacro\compA{1}      %Component R or H
  \pgfmathsetmacro\compB{#1/2000} %Component G or S
  \pgfmathsetmacro\compC{1}      %Component B or B
  \edef\x{\noexpand\centering\noexpand\cellcolor[\colorModel]{\compA,\compB,\compC}}\x #1
  }
\newcommand\ColCellP[1]{
  \pgfmathparse{#1<5000?1:0}  %Threshold for changing the font color into the cells
    \ifnum\pgfmathresult=0\relax\color{white}\fi
  \pgfmathsetmacro\compA{1}      %Component R or H
  \pgfmathsetmacro\compB{#1/200} %Component G or S
  \pgfmathsetmacro\compC{1}      %Component B or B
  \edef\x{\noexpand\centering\noexpand\cellcolor[\colorModel]{\compA,\compB,\compC}}\x #1
  }
\newcommand\ColCellPP[1]{
  \pgfmathparse{#1<5000?1:0}  %Threshold for changing the font color into the cells
    \ifnum\pgfmathresult=0\relax\color{white}\fi
  \pgfmathsetmacro\compA{1}      %Component R or H
  \pgfmathsetmacro\compB{#1/1.2} %Component G or S
  \pgfmathsetmacro\compC{1}      %Component B or B
  \edef\x{\noexpand\centering\noexpand\cellcolor[\colorModel]{\compA,\compB,\compC}}\x #1
  }
\newcommand\ColCellPPP[1]{
  \pgfmathparse{#1<5000?1:0}  %Threshold for changing the font color into the cells
    \ifnum\pgfmathresult=0\relax\color{white}\fi
  \pgfmathsetmacro\compA{1}      %Component R or H
  \pgfmathsetmacro\compB{#1} %Component G or S
  \pgfmathsetmacro\compC{1}      %Component B or B
  \edef\x{\noexpand\centering\noexpand\cellcolor[\colorModel]{\compA,\compB,\compC}}\x #1
  }
\newcolumntype{E}{>{\collectcell\ColCell}m{2.2em}<{\endcollectcell}}
\newcolumntype{U}{>{\collectcell\ColCellP}m{2.1em}<{\endcollectcell}}
\newcolumntype{F}{>{\collectcell\ColCellPP}m{2.2em}<{\endcollectcell}}
\newcolumntype{G}{>{\collectcell\ColCellPPP}m{2.1em}<{\endcollectcell}}

\title{SICKNL: A Dataset for Dutch Natural Language Inference}
\author{Gijs Wijnholds \\ UiL-OTS \\ Utrecht University \\ \texttt{g.j.wijnholds@uu.nl} \\\And
Michael Moortgat \\ UiL-OTS \\ Utrecht University \\
\texttt{m.j.moortgat@uu.nl}}

\date{\today}

\begin{document}
\maketitle
\begin{abstract}
We present SICK-NL (read: \emph{signal}), a dataset targeting Natural Language Inference in Dutch. SICK-NL is obtained by translating the SICK dataset of \citet{marelli-etal-2014-sick} from English into Dutch. Having a parallel inference dataset allows us to compare both monolingual and multilingual NLP models for English and Dutch on the two tasks.
In the paper, we motivate and detail the translation process, perform a baseline evaluation on both the original SICK dataset and its Dutch incarnation SICK-NL, taking inspiration from Dutch skipgram embeddings and contextualised embedding models. In addition, we encapsulate two phenomena encountered in the translation to formulate \emph{stress tests} and verify how well the Dutch models capture syntactic restructurings that do not affect semantics. Our main finding is all models perform worse on SICK-NL than on SICK, indicating that the Dutch dataset is more challenging than the English original. Results on the stress tests show that models don't fully capture word order freedom in Dutch, warranting future systematic studies.
\end{abstract}

\section{Introduction}

One of the primary tasks for Natural Language Processing (NLP) systems is Natural Language Inference (NLI), where the goal is to determine, for a given premise sentence whether it contradicts, entails, or is neutral with respect to a given hypothesis sentence. 

For English, several standard NLI datasets exist, such as SICK \cite{marelli-etal-2014-sick}, SNLI \cite{bowman-etal-2015-large} and MNLI \cite{williams-etal-2018-broad}. Having such inference datasets available only for English may introduces a bias in NLP research. \citet{conneau-etal-2018-xnli} introduce XNLI, a multilingual version of a fragment of the SNLI dataset, that contains pairs for Natural Language Inference in 15 languages and is explicitly intended to serve as a resource for evaluating crosslingual representations. However, Dutch is not represented in any current NLI dataset, a lack that we wish to complement.
% For English, several standard NLI datasets exist, such as SICK (Sentences Involving Compositional Knowledge, \cite{marelli-etal-2014-sick}), SNLI (Stanford Natural Language Inference, \cite{bowman-etal-2015-large}) and MNLI (Multi-genre Natural Language Inference, \cite{williams-etal-2018-broad}). Having such inference datasets available only for English may introduces a bias in NLP research. \citet{williams2017broad} introduce XNLI, a multilingual version of a fragment of the SNLI dataset, that contains pairs for Natural Language Inference in 15 languages and is explicitly intended to serve as a resource for evaluating crosslingual representations.
% We wish to complement the lack of Natural Language Inference resources for Dutch. 

Dutch counts as a high-resource language, with the sixth largest Wikipedia (2M+ articles), despite having ca. 25M native speakers. Moreover, the syntactically parsed LASSY corpus of written Dutch \cite{vanNoord2013}, and the SONAR corpus of written Dutch \cite{Oostdijk2013} provide rich resources on which NLP systems may be developed. Indeed, Dutch is in the scope of the multilingual BERT models published by Google \cite{devlin-etal-2019-bert}, and two monolingual Dutch BERT models have been published as part of HuggingFace's transformers library \cite{de2019bertje,delobelle2020robbert}.

Compared to English, however, the number of evaluation tasks for Dutch is limited. There is a Named Entity Recognition task coming from the CoNLL-2003 shared task \cite{tjong2003introduction}; from a one million word hand annotated subcorpus of SONAR \cite{Oostdijk2013} one derives part-of-speech tagging, Named Entity Recognition and Semantic Role Labelling tasks. More recently a Sentiment Analysis dataset was introduced, based on Dutch Book reviews \cite{van2019merits}. Moreover, \citet{allein2020binary} introduce a classification task where a model needs to distinguish between the pronouns \emph{die} and \emph{dat}.

Given the focus on word/token-level tasks in Dutch, we aim to complement existing resources with an NLI task for Dutch. We do so by deriving it from the English SICK dataset, for the following reasons: first, this dataset requires a small amount of world knowledge as it was derived mainly from image captions that are typically concrete descriptions of a scene. Therefore, no world knowledge requirements will be imposed on an NLP model for the task, but rather its ability for reasoning will be assessed. Secondly, due to the structure of the sentences in SICK, the types of inferences can be  attributed to particular constructs, such as hypernymy/hyponymy, negation, or choice of quantification. Thirdly, SICK contains 6076 unique sentences and almost 10K inference pairs, making it a sizeable dataset for NLP standards, while deriving a Dutch version is more manageable than with other datasets. We make the dataset, code and derived resources (see Section \ref{section:stresstests}), available online\footnote{\href{https://github.com/gijswijnholds/sick_nl}{\nolinkurl{github.com/gijswijnholds/sick_nl}}}.

\section{Dataset Creation}

We follow a semi-automatic translation procedure to create SICK-NL, similar to the Portuguese version of SICK \cite{real2018sick}. First, we use a machine translator to translate all of the (6076) unique sentences of SICK. We review each sentence and its translation in parallel, correcting any mistakes made by the machine translator, and maintaining consistency of individual words' translation, in the process guaranteeing that the meaning of each sentence is preserved as much as possible. Finally, we perform a postprocessing step in which we ensure unique translations for unique sentences (alignment) with as few exceptions as possible. In this way we obtain 6059 unique Dutch sentences, which means that the dataset is almost fully aligned on the sentence level. It should be noted, however, that can not fully guarantee the same choice of words in each sentence \emph{pair} in the original dataset, as we translate sentence by sentence.

Table \ref{table:statistics} shows some statistics of SICK and its Dutch translation. The most notable difference is that the amount of unique words in Dutch is 23\% higher than that in English, even though the total number of words in SICK-NL is about 93\% of that in SICK. We argue that this is due to morphological complexities of Dutch, where verbs can be compound and be split up into multiple parts (for example, ``\emph{storing}" becomes \emph{``opbergen"}, which may be used as ``\emph{de man bergt iets op}"). Moreover, Dutch enjoys a relatively free word order, which in the case of SICK-NL means that sometimes the order of the main verb and its direct object may be swapped in the sentence, especially when the present continuous form (``\emph{is cutting an onion}") is preserved in Dutch (``\emph{is een ui aan het snijden}"). Finally, we follow the machine translation, only making changes in the case of grammatical errors, lexical choice inconsistencies, and changes in meaning. This freedom leads to a decrease in relative word overlap between premise and hypothesis sentence, computed as the number of words in common divided by the length of the shortest sentence. From the perspective of Natural Language Inference this is preferable as word overlap often can be exploited by neural network architectures \cite{mccoy-etal-2019-right}.

% After translation, we want to find out to what extent the Dutch translation of SICK reflects the statistical properties of the original dataset, or whether these have been overwritten by typicalities of the Dutch language or choices made in translation. 

% Specific constructions due to the relative free word order in Dutch may complicate the Dutch dataset over the English original.
% Dutch enjoys a relatively free word order, which in the case of SICK-NL means that the same sentence can often be described in various different ways in Dutch, where notably the order of verb and direct object in a sentence can be swapped. Moreover, Dutch morphology is closer to German than it is to English, and notably compound words are built up by concatenation, rather than juxtaposition as it is done in English. For example, the word ``\emph{race driver}" will be translated by ``\emph{racecoureur}".

\begin{table}[]
\centering
    \begin{tabular}{@{}lcc@{}}
    \toprule
            & \textbf{SICK} & \textbf{SICK-NL} \\
    \midrule
    No. of tokens        & 189783 & 176509 \\
    No. of unique tokens & 2328  & 2870 \\
    Avg. sentence length & 9.64 & 8.97 \\
    Avg. word overlap & 66.91\% & 58.99\% \\
    \bottomrule
    \end{tabular}
    \caption{Basic statistics of SICK and SICK-NL.}
    \label{table:statistics}
    \vspace{-1em}
\end{table}

\section{Baseline Evaluation and Results}

We evaluate two types of models as a baseline to compare SICK-NL with its English original.

First, we evaluate embeddings that were not specifically trained on SICK. Table \ref{table:baseline_sick_models} shows the correlation results on the relatedness task of SICK and SICK-NL, where the cosine similarity between two independently computed sentence embeddings is correlated with the relatedness scores of human annotators (between 1 and 5).

\begin{table}[h!]
    \begin{tabular}{@{}lcl@{\hskip 0.3em}c@{}}
        \toprule
         & \textbf{SICK} & & \textbf{SICK-NL} \\
         \midrule
         Skipgram & 69.49 & Skipgram & 56.94 \\
         BERT$_{cls}$ & 50.78 & BERTje$_{cls}$ & 49.06 \\
         BERT$_{avg}$ & 61.36 & BERTje$_{avg}$ & 55.55 \\
         RoBERTa$_{cls}$ & 46.62 & RobBERT$_{cls}$ & 43.93 \\
         RoBERTa$_{avg}$ & 62.71 & RobBERT$_{avg}$ & 52.33 \\
         \bottomrule
    \end{tabular}
    \caption{Pearson $r$ correlation coefficient for the relatedness task of the English SICK dataset (left) and its Dutch translation (right).}
    \label{table:baseline_sick_models}
    \vspace{-1em}
\end{table}
To obtain sentence embeddings here, we average skipgram embeddings, or, in the case of contextualised  embeddings, we take either the sentence embedding given by the $[CLS]$ token, or we take the average of the individual word's embeddings. For the skipgram embeddings in English, we use the standard 300-dimensional GoogleNews vectors provided by the \verb|word2vec| package and for Dutch, we use the 320-dimensional Wikipedia trained embeddings of \citet{tulkens-etal-2016-evaluating}.

The relatedness results show that (a) using the $[CLS]$ token embedding as a sentence encoding performs worse than taking the average of word embeddings, and that (b) the Dutch incarnation of SICK is harder than the original English dataset.

In the second setup, we use BERTje, the Dutch BERT model of \citet{de2019bertje} and RobBERT, the Dutch RoBERTa model of \citet{delobelle2020robbert}, with their corresponding English counterparts, as well as multilingual BERT (mBERT), as sequence classifiers on the Entailment task of SICK(-NL). Here we observe a similar pattern in the results in Table \ref{table:bert_finetune_entailment}: while there are individual difference on the same task, the main surprise is that the Dutch dataset is harder, even when exactly the same model (mBERT) is used.

\begin{table}[h!]
\begin{center}
    \begin{tabular}{@{}lclc@{}}
        \toprule
         & \textbf{SICK} & & \textbf{SICK-NL} \\
         \midrule
         BERT & 87.34 & BERTje & 83.94 \\
         mBERT & 87.02 & mBERT & 84.53 \\
         RoBERTa & 90.11 & RobBERT & 82.02 \\
         \bottomrule
        %  12 & 9 \\
        %  15 & 20 (o.w. we can take 7 as well) \\
        %   5 & 10 \\
    \end{tabular}
\end{center}
    \caption{Accuracy results on the entailment task of the English SICK dataset and its Dutch translation for two Dutch BERT models and their English counterparts. For each model, we report the best score out of 20 epochs of fine-tuning.}
    \label{table:bert_finetune_entailment}
    \vspace{-1em}
\end{table}

\section{Error Analysis}

In order to understand the differences between the Dutch and English language models on the respective tasks, we dive deeper into the classification results. We plot confusion matrices for each model in Table \ref{table:confusion_matricesbert}, where we separate predictions that the models have in common and the from the predictions that are unique to each model.

In the case of English, performance on classifying contradictions is worse for multilingual BERT and RoBERTa, and RoBERTa also gives highest recall values for the Neutral and Entailment labels. This is all not surprising given that RoBERTa has the overall highest test set accuracy. The surprising results come mainly from the comparison between English and Dutch models. Where BERTje is rather indecisive when it comes to Neutral sentence pairs (it classifies roughly equal numbers as Neutral and Entailment), it classifies 74\% of Entailment pairs as Neutral. For multilingual BERT the situation is reversed, with 47\% of Neutral entailments classified as Entailment, although for cases of entailment, the classifier did not clearly distinguish Neutral from Entailment. The most surprising pattern was observed in RobBERT: where RoBERTa still has high recall for Neutral and Entailment, its Dutch counterpart RobBERT mistakes most Neutral cases as Entailment and even more so vice versa. For all models, in these four cases of misclassification, in the case of the English task the correct inference was made in at least 99\% of the cases.

Following \citet{naik-etal-2018-stress}, we inspect these prominent cases of misclassification in Dutch by looking at the number of cases of high overlap (at most four words not in common), and at the number of length mismatches (the difference between sentence length exceeds 4), and set off these distributions against that of the test set, in Table \ref{table:error_statistics}. \vspace{-1em}
% We also checked for negation (the words `niet' and `geen') and the presence of a present continuous form, but found no significant distributions.
\begin{table}[h!]
\centering
    \begin{tabular}{@{}l@{\hskip 1em}c@{\hskip 0.75em}c@{\hskip 0.75em}c@{\hskip 0.75em}c@{\hskip 1em}c@{}}
    \toprule
    % & pass. & word diff. & len diff. & neg. & rest \\
    & \footnotesize{\textbf{BERT}} &
    \footnotesize{\textbf{mBERT}} &
    \footnotesize{\textbf{RobBERT}} & \footnotesize{\textbf{RobBERT}} & \footnotesize{\textbf{Test}} \\
    & \footnotesize{(N$\rightarrow$E)} & \footnotesize{(E$\rightarrow$N)} & \footnotesize{(N$\rightarrow$E)} & \footnotesize{(E$\rightarrow$N)} & \\
    \midrule
    & \multicolumn{4}{c}{Word difference} & \\
    \midrule
    EN &  66\% & 42\% & 62\% & 44\% & 40\% \\
    NL & 47\% & 38\% & 41\% & 38\% & 28\% \\
    \midrule
    & \multicolumn{4}{c}{Length mismatch} & \\
    \midrule
    EN & 24 \% & 17\% & 25\% & 17\% & 27\% \\
    NL & 30 \% & 31\% & 28\% & 25\% & 36\% \\
    %  .66 & .42 & .62 & .44 & .40 \\
    %  .47 & .38 & .41 & .38 & .28 \\
    %  .24  & .17 & .25 & .17 & .27 \\
    %  .30  & .31 & .28 & .25 & .36 \\
    % \small{word diff. EN}   & 66\% & 42\% & 62\% & 44\% & 40\% \\
    % \small{word diff. NL}   & 47\% & 38\% & 41\% & 38\% & 28\% \\
    % \small{len diff. EN} & 24 \% & 17\% & 25\% & 17\% & 27\% \\
    % \small{len diff. NL} & 30 \% & 31\% & 28\% & 25\% & 36\% \\
    \bottomrule
    \end{tabular}
    \caption{Error analysis of prominent misclassifications.}
    \label{table:error_statistics}
    \vspace{-0.5em}
\end{table}

The main finding here is that word overlap does provide a strong cue in the English dataset, especially given that SICK has more cases (1970) overall than SICK-NL (1385), and that in SICK they are more concentrated in cases of Entailment. Length mismatches occur more often in SICK-NL but seem to provide less of a cue to the models to make strong inference decisions.

\begin{table*}[]
    \hspace{-1.1em}\begin{tabular}{l@{\hskip 0.2em}c@{\hskip 0.2em}c}
    \begin{tabular}{clEEEc}
    \multicolumn{2}{c}{\textbf{\small{BERT}}} & \multicolumn{3}{c}{\textbf{Prediction EN-NL}} \\
    & & \multicolumn{1}{c}{\textbf{C}} & \multicolumn{1}{c}{\textbf{N}} & \multicolumn{1}{c}{\textbf{E}} & \emph{rec.} \\
    \multirow{3}{*}{\rotatebox{90}{\textbf{Gold}}}
    & \textbf{C} & 549 & 62 & 16 & 88\% \\
    & \textbf{N} & 35 & 2341 & 147 & 93\% \\
    & \textbf{E} & 3 & 143 & 1035 & 88\% \\
    & \emph{pr.} & \multicolumn{1}{c}{94\%} & \multicolumn{1}{c}{92\%} & \multicolumn{1}{c}{86\%} & \\
    \end{tabular}
    &
    \begin{tabular}{UUUc}
    \multicolumn{3}{c}{\textbf{Prediction EN}} \\
    \multicolumn{1}{c}{\textbf{C}} & \multicolumn{1}{c}{\textbf{N}} & \multicolumn{1}{c}{\textbf{E}} & \emph{rec.} \\
    46 & 27 & 12 & 54\% \\
    37 & 147 & 83 & 55\% \\
    3 & 53 & 167 & 75\% \\
    \multicolumn{1}{c}{53\%} & \multicolumn{1}{c}{65\%} & \multicolumn{1}{c}{64\%} & \\
    \end{tabular}
    &
    \begin{tabular}{UUUc}
    \multicolumn{3}{c}{\textbf{Prediction NL}} \\
    \multicolumn{1}{c}{\textbf{C}} & \multicolumn{1}{c}{\textbf{N}} & \multicolumn{1}{c}{\textbf{E}} & \multicolumn{1}{c}{\emph{rec.}} \\
    22 & 52 & 11 & 26\% \\
    32 & 116 & 119 & 43\% \\
    3 & 165 & 55 & 25\% \\
    \multicolumn{1}{c}{39\%} & \multicolumn{1}{c}{35\%} & \multicolumn{1}{c}{30\%} & \\
    \end{tabular} \\
    % \vspace{-1em} & & \\
    \begin{tabular}{clEEEc}
    \multicolumn{2}{c}{\textbf{\small{mBERT}}} \\
    \multirow{3}{*}{\rotatebox{90}{\textbf{Gold}}}
    & \textbf{C} & 553 & 69 & 10 & 88\% \\
    & \textbf{N} & 32 & 2344 & 106 & 94\% \\
    & \textbf{E} & 3 & 160 & 1015 & 86\% \\
    & \emph{pr.} &\multicolumn{1}{c}{94\%} & \multicolumn{1}{c}{91\%} & \multicolumn{1}{c}{90\%} & \\
    \end{tabular}
    &
    \begin{tabular}{UUUc}
    \multicolumn{3}{c}{} \\
    24 & 47 & 9 & 30\% \\
    9 & 210 & 89 & 68\% \\
    0 & 103 & 123 & 54\% \\
    \multicolumn{1}{c}{73\%} & \multicolumn{1}{c}{58\%} & \multicolumn{1}{c}{56\%} & \\
    \end{tabular}
    &
    \begin{tabular}{UUUc}
    \multicolumn{3}{c}{} \\
    42 & 31 & 7 & 52\% \\
    71 & 93 & 144 & 30\% \\
    15 & 111 & 100 & 44\% \\
    \multicolumn{1}{c}{33\%} & \multicolumn{1}{c}{40\%} & \multicolumn{1}{c}{40\%} & \\
    \end{tabular} \\
    % \vspace{-1em} & & \\
    \begin{tabular}{clEEEc}
    \multicolumn{2}{c}{\textbf{\small{RoBERTa}}} \\
    \multirow{3}{*}{\rotatebox{90}{\textbf{Gold}}}
    & \textbf{C} & 563 & 68 & 4 & 89\% \\
    & \textbf{N} & 25 & 2301 & 82 & 96\% \\
    & \textbf{E} & 1 & 108 & 995 & 90\% \\
    & \emph{pr.} &\multicolumn{1}{c}{96\%} & \multicolumn{1}{c}{93\%} & \multicolumn{1}{c}{92\%} & \\
    \end{tabular}
    &
    \begin{tabular}{UUUc}
    \multicolumn{3}{c}{} \\
    27 & 46 & 4 & 35\% \\
    6 & 279 & 97 & 73\% \\
    2 & 42 & 256 & 85\% \\
    \multicolumn{1}{c}{77\%} & \multicolumn{1}{c}{76\%} & \multicolumn{1}{c}{72\%} & \\
    \end{tabular}
    &
    \begin{tabular}{UUUc}
    \multicolumn{3}{c}{} \\
    35 & 28 & 14 & 45\% \\
    79 & 96 & 207 & 25\% \\
    15 & 246 & 39 & 13\% \\
    \multicolumn{1}{c}{27\%} & \multicolumn{1}{c}{26\%} & \multicolumn{1}{c}{15\%} & \\
    \end{tabular}
    \end{tabular}
    \caption{Confusion matrices for English vs Dutch language models, finetuned. Top: BERT vs BERTje.  The models disagree in 13.3\% of cases. Middle: Multilingual BERT. The model disagrees in 14.3\% of cases). Bottom: Roberta vs RobBERT. The models (disagree in 18.3\% of cases).}
    % \caption{Confusion matrices for English vs Dutch language models, finetuned. Top: BERT vs BERTje.  The models agree on 86.7\% of the cases (disagree in 13.3\% of cases). Middle: Multilingual BERT. The model agrees on the two datasets in 85.7\% of cases (disagrees in 14.3\% of cases). Bottom: Roberta vs RobBERT. The models agree on 81.7\% of the cases (disagree in 18.3\% of cases).}
    \label{table:confusion_matricesbert}
    \vspace{-1em}
\end{table*}

\section{Stress Testing}
\label{section:stresstests}
One of the potential sources of error could have been the passive form translation of a verb. Such constructions, combined with a prepositional phrase, form an interesting testbed for Dutch as they allow the prepositional phrase to be moved in front of the verb in a sentence without changing the meaning. For example, ``\emph{Een vrouw is aan het wakeboarden op een meer}" (``\emph{A woman is wakeboarding on a lake}"), may in Dutch be used interchangeably with ``\emph{Een vrouw is op een meer aan het wakeboarden}"). We select all (88) sentences in SICK-NL that contain both the `aan het' construction and a prepositional phrase, and generate their permutations. Then, we replace all (225) inference pairs with these sentences such that they now contain a sentence with different word order but the exact same meaning and therefore the inference label is preserved. We then verify how the model's predictions do on those inference pairs that were in the test set (116). Additionally, we check whether the models are able to interchange sentences and their rewritten equivalent (i.e. classify as Entailment).

As a second test, we investigate the role of the simple present versus the present continuous. We take all the (383) cases of present continuous in the Dutch dataset and replace them by a simple present equivalent, leading to 1137 pairs, out of which 576 occur in the test data. For example, we turn the sentence ``\emph{De man is aan het zwemmen}" into the simple form ``\emph{De man zwemt}". We then repeat the same procedure as above, asking how many inference predictions change as a result of this form change, and whether the forms can be used interchangeably for the models.
\begin{table}[h]
\vspace{0.5em}
\centering
    \begin{tabular}{@{}lcccc@{}}
    \toprule
        & \multicolumn{4}{c}{present cont. $\rightarrow$ present simple} \\
        & Before & After & $\rightarrow$ & $\leftarrow$ \\
    \midrule
    \textbf{BERT}    & 84.58 & 86.66 & 93.21 & 92.43 \\
    \textbf{mBERT}  & 86.14 & 84.92 & 94.26 & 94.52\\
    \textbf{RobBERT} & 82.84 & 81.98 & 86.16 & 84.33\\
    \midrule
        & \multicolumn{4}{c}{prep. phrase order switch} \\
    \midrule
    \textbf{BERT}    & 81.03 & 78.45 & 85.23 & 85.23 \\
    \textbf{mBERT}  & 87.93 & 85.34 & 85.23 & 80.68 \\
    \textbf{RobBERT} & 76.72 & 75.86 & 72.73 & 73.86 \\
    \bottomrule
    \end{tabular}
    \caption{Stress test accuracy. Left: accuracy before and after rewriting. Right: inference between rewritings.}
    \label{table:stresstest_results}
    \vspace{-1em}
\end{table}

The results in Table \ref{table:stresstest_results}  indicate that the interchange between present continuous and simple present forms does not make much of a difference to the models' performance, and interchangeability is high except for RobBERT that scores under 90\%. However, switching the order of prepositional phrase and verb has a much stronger effect with all models consistently scoring lower on the relevant part of the test set, and mainly the models being particularly poor at interchanging these sentences that are semantically equivalent.

% Negative findings: the difference was not caused by:
%     \begin{enumerate}
%         \item Translating infinitival verb constructions into infinitival verb constructions in Dutch. Many of the English examples have the form ".... is ... VERBing ....", which can be translated as either ".... is ... aan het ... VERBen ...", or as the direct form "..... VERBt ...". Inspection shows that the distribution of these translations are the same among the overlapping and unique examples.
%         \item Word overlap. We did not find any significant difference in word overlap among misclassification categories.
%         \item Vocabulary difference. There was no observable significant difference between vocabulary size between overlapping and unique predictions.
%     \end{enumerate}
\section{Conclusion}
In this paper we introduced an NLI dataset for Dutch by semi-automatically translating the SICK dataset. To our knowledge this is the first available inference task for Dutch. Despite the common perception that Dutch is very similar to English, SICK-NL was significantly more difficult to tackle, even for language models that had access to the training data for fine-tuning. We hypothesised that the difference in result may be due to a larger vocabulary in SICK-NL, and a decline in word overlap between inference pairs. In addition we performed two stress tests and found that pretrained models that were exposed to the training data had difficulty detecting semantically equivalent sentences that differ only in word order. Further work will therefore more systematically assess such phenomena.
\bibliography{references}

\begin{thebibliography}{16}
\expandafter\ifx\csname natexlab\endcsname\relax\def\natexlab#1{#1}\fi

\bibitem[{Allein et~al.(2020)Allein, Leeuwenberg, and Moens}]{allein2020binary}
Liesbeth Allein, Artuur Leeuwenberg, and Marie-Francine Moens. 2020.
\newblock \href {https://arxiv.org/pdf/2001.02943.pdf} {Binary and multitask
  classification model for {D}utch anaphora resolution: Die/dat prediction}.
\newblock \emph{arXiv preprint arXiv:2001.02943}.

\bibitem[{Bowman et~al.(2015)Bowman, Angeli, Potts, and
  Manning}]{bowman-etal-2015-large}
Samuel~R. Bowman, Gabor Angeli, Christopher Potts, and Christopher~D. Manning.
  2015.
\newblock \href {https://doi.org/10.18653/v1/D15-1075} {A large annotated
  corpus for learning natural language inference}.
\newblock In \emph{Proceedings of the 2015 Conference on Empirical Methods in
  Natural Language Processing}, pages 632--642, Lisbon, Portugal. Association
  for Computational Linguistics.

\bibitem[{van~der Burgh and Verberne(2019)}]{van2019merits}
Benjamin van~der Burgh and Suzan Verberne. 2019.
\newblock \href {https://arxiv.org/pdf/1910.00896.pdf} {The merits of universal
  language model fine-tuning for small datasets--a case with {D}utch book
  reviews}.
\newblock \emph{arXiv preprint arXiv:1910.00896}.

\bibitem[{Conneau et~al.(2018)Conneau, Rinott, Lample, Williams, Bowman,
  Schwenk, and Stoyanov}]{conneau-etal-2018-xnli}
Alexis Conneau, Ruty Rinott, Guillaume Lample, Adina Williams, Samuel Bowman,
  Holger Schwenk, and Veselin Stoyanov. 2018.
\newblock \href {https://doi.org/10.18653/v1/D18-1269} {{XNLI}: Evaluating
  cross-lingual sentence representations}.
\newblock In \emph{Proceedings of the 2018 Conference on Empirical Methods in
  Natural Language Processing}, pages 2475--2485, Brussels, Belgium.
  Association for Computational Linguistics.

\bibitem[{Delobelle et~al.(2020)Delobelle, Winters, and
  Berendt}]{delobelle2020robbert}
Pieter Delobelle, Thomas Winters, and Bettina Berendt. 2020.
\newblock \href {https://arxiv.org/pdf/2001.06286.pdf} {Rob{BERT}: a {D}utch
  {R}o{BERT}a-based language model}.
\newblock \emph{arXiv preprint arXiv:2001.06286}.

\bibitem[{Devlin et~al.(2019)Devlin, Chang, Lee, and
  Toutanova}]{devlin-etal-2019-bert}
Jacob Devlin, Ming-Wei Chang, Kenton Lee, and Kristina Toutanova. 2019.
\newblock \href {https://doi.org/10.18653/v1/N19-1423} {{BERT}: Pre-training of
  deep bidirectional transformers for language understanding}.
\newblock In \emph{Proceedings of the 2019 Conference of the North {A}merican
  Chapter of the Association for Computational Linguistics: Human Language
  Technologies, Volume 1 (Long and Short Papers)}, pages 4171--4186,
  Minneapolis, Minnesota. Association for Computational Linguistics.

\bibitem[{Marelli et~al.(2014)Marelli, Menini, Baroni, Bentivogli, Bernardi,
  and Zamparelli}]{marelli-etal-2014-sick}
Marco Marelli, Stefano Menini, Marco Baroni, Luisa Bentivogli, Raffaella
  Bernardi, and Roberto Zamparelli. 2014.
\newblock \href
  {http://www.lrec-conf.org/proceedings/lrec2014/pdf/363_Paper.pdf} {A {SICK}
  cure for the evaluation of compositional distributional semantic models}.
\newblock In \emph{Proceedings of the Ninth International Conference on
  Language Resources and Evaluation ({LREC}-2014)}, pages 216--223, Reykjavik,
  Iceland. European Languages Resources Association (ELRA).

\bibitem[{McCoy et~al.(2019)McCoy, Pavlick, and Linzen}]{mccoy-etal-2019-right}
Tom McCoy, Ellie Pavlick, and Tal Linzen. 2019.
\newblock \href {https://doi.org/10.18653/v1/P19-1334} {Right for the wrong
  reasons: Diagnosing syntactic heuristics in natural language inference}.
\newblock In \emph{Proceedings of the 57th Annual Meeting of the Association
  for Computational Linguistics}, pages 3428--3448, Florence, Italy.
  Association for Computational Linguistics.

\bibitem[{Naik et~al.(2018)Naik, Ravichander, Sadeh, Rose, and
  Neubig}]{naik-etal-2018-stress}
Aakanksha Naik, Abhilasha Ravichander, Norman Sadeh, Carolyn Rose, and Graham
  Neubig. 2018.
\newblock \href {https://www.aclweb.org/anthology/C18-1198} {Stress test
  evaluation for natural language inference}.
\newblock In \emph{Proceedings of the 27th International Conference on
  Computational Linguistics}, pages 2340--2353, Santa Fe, New Mexico, USA.
  Association for Computational Linguistics.

\bibitem[{van Noord et~al.(2013)van Noord, Bouma, Van~Eynde, de~Kok, van~der
  Linde, Schuurman, Sang, and Vandeghinste}]{vanNoord2013}
Gertjan van Noord, Gosse Bouma, Frank Van~Eynde, Dani{\"e}l de~Kok, Jelmer
  van~der Linde, Ineke Schuurman, Erik Tjong~Kim Sang, and Vincent
  Vandeghinste. 2013.
\newblock \href {https://doi.org/10.1007/978-3-642-30910-6_9} {\emph{Large
  Scale Syntactic Annotation of Written Dutch: Lassy}}, pages 147--164.
  Springer Berlin Heidelberg, Berlin, Heidelberg.

\bibitem[{Oostdijk et~al.(2013)Oostdijk, Reynaert, Hoste, and
  Schuurman}]{Oostdijk2013}
Nelleke Oostdijk, Martin Reynaert, V{\'e}ronique Hoste, and Ineke Schuurman.
  2013.
\newblock \href {https://doi.org/10.1007/978-3-642-30910-6_13} {\emph{The
  Construction of a 500-Million-Word Reference Corpus of Contemporary Written
  Dutch}}, pages 219--247. Springer Berlin Heidelberg, Berlin, Heidelberg.

\bibitem[{Real et~al.(2018)Real, Rodrigues, Vieira~e Silva, Albiero,
  Thalenberg, Guide, Silva, de~Oliveira~Lima, C{\^a}mara, Stanojevi{\'{c}},
  Souza, and de~Paiva}]{real2018sick}
Livy Real, Ana Rodrigues, Andressa Vieira~e Silva, Beatriz Albiero, Bruna
  Thalenberg, Bruno Guide, Cindy Silva, Guilherme de~Oliveira~Lima, Igor C.~S.
  C{\^a}mara, Milo{\v{s}} Stanojevi{\'{c}}, Rodrigo Souza, and Valeria
  de~Paiva. 2018.
\newblock \href
  {https://link.springer.com/chapter/10.1007/978-3-319-99722-3_31} {{SICK-BR}:
  A {P}ortuguese corpus for inference}.
\newblock In \emph{Computational Processing of the Portuguese Language}, pages
  303--312, Cham. Springer International Publishing.

\bibitem[{Tjong Kim~Sang and De~Meulder(2003)}]{tjong2003introduction}
Erik~F. Tjong Kim~Sang and Fien De~Meulder. 2003.
\newblock \href {https://doi.org/10.3115/1119176.1119195} {Introduction to the
  {C}o{NLL}-2003 shared task: Language-independent named entity recognition}.
\newblock In \emph{Proceedings of the Seventh Conference on Natural Language
  Learning at HLT-NAACL 2003 - Volume 4}, CONLL '03, page 142–147, USA.
  Association for Computational Linguistics.

\bibitem[{Tulkens et~al.(2016)Tulkens, Emmery, and
  Daelemans}]{tulkens-etal-2016-evaluating}
St{\'e}phan Tulkens, Chris Emmery, and Walter Daelemans. 2016.
\newblock \href {https://www.aclweb.org/anthology/L16-1652} {Evaluating
  unsupervised {D}utch word embeddings as a linguistic resource}.
\newblock In \emph{Proceedings of the Tenth International Conference on
  Language Resources and Evaluation ({LREC}'16)}, pages 4130--4136,
  Portoro{\v{z}}, Slovenia. European Language Resources Association (ELRA).

\bibitem[{de~Vries et~al.(2019)de~Vries, van Cranenburgh, Bisazza, Caselli, van
  Noord, and Nissim}]{de2019bertje}
Wietse de~Vries, Andreas van Cranenburgh, Arianna Bisazza, Tommaso Caselli,
  Gertjan van Noord, and Malvina Nissim. 2019.
\newblock \href {https://arxiv.org/pdf/1912.09582.pdf} {{BERT}je: A {D}utch
  {BERT} model}.
\newblock \emph{arXiv preprint arXiv:1912.09582}.

\bibitem[{Williams et~al.(2018)Williams, Nangia, and
  Bowman}]{williams-etal-2018-broad}
Adina Williams, Nikita Nangia, and Samuel Bowman. 2018.
\newblock \href {https://doi.org/10.18653/v1/N18-1101} {A broad-coverage
  challenge corpus for sentence understanding through inference}.
\newblock In \emph{Proceedings of the 2018 Conference of the North {A}merican
  Chapter of the Association for Computational Linguistics: Human Language
  Technologies, Volume 1 (Long Papers)}, pages 1112--1122, New Orleans,
  Louisiana. Association for Computational Linguistics.

\end{thebibliography}
\bibliographystyle{acl_natbib}

% \bibliography{anthology,eacl2021}
% \bibliographystyle{acl_natbib}

\end{document}